\documentclass[conference]{IEEEtran}
\IEEEoverridecommandlockouts
\usepackage{cite}
\usepackage{amsmath,amssymb,amsfonts}
\usepackage{algorithmic}
\usepackage{pifont}
\newcommand{\cmark}{\ding{51}} 
\newcommand{\xmark}{\ding{55}} 
\usepackage{algorithm}
\usepackage{graphicx}
\usepackage{textcomp}
\usepackage{xcolor}
\usepackage{caption}
\usepackage{comment}
\usepackage{tikz}
\usepackage{tikz-3dplot}
\usepackage{booktabs}
\usepackage{tabularx} 
\usepackage[colorlinks=true, linkcolor=black, citecolor=black, urlcolor=black]{hyperref}

\usepackage{xurl}          
\usetikzlibrary{arrows.meta, positioning}

\def\BibTeX{{\rm B\kern-.05em{\sc i\kern-.025em b}\kern-.08em
    T\kern-.1667em\lower.7ex\hbox{E}\kern-.125emX}}

\begin{document}

\title{\textbf{RPGD}: \textbf{R}ANSAC-\textbf{P}3P \textbf{G}radient \textbf{D}escent for Extrinsic Calibration in 3D Human Pose Estimation\\

{\footnotesize }

\thanks{This work is co-funded by the European Union’s Horizon Europe research and innovation programme under Marie Skłodowska-Curie Actions (MSCA) with grant agreement No 101081674.} 

}

\author{\IEEEauthorblockN{Zhanyu Tuo}
\IEEEauthorblockA{
\textit{Sorbonne university}\\
Paris, France \\
zhanyu.tuo@sorbonne-universite.fr} 
}

\maketitle
\begin{abstract}

In this paper, we propose RPGD (RANSAC-P3P Gradient Descent), a human-pose-driven extrinsic calibration framework that robustly aligns MoCap-based 3D skeletal data with monocular or multi-view RGB cameras using only natural human motion. RPGD formulates extrinsic calibration as a coarse-to-fine problem tailored to human poses, combining the global robustness of RANSAC-P3P with Gradient-Descent-based refinement. We evaluate RPGD on three large-scale public 3D HPE datasets as well as on a self-collected in-the-wild dataset. Experimental results demonstrate that RPGD consistently recovers extrinsic parameters with accuracy comparable to the provided ground truth, achieving sub-pixel MPJPE reprojection error even in challenging, noisy settings. These results indicate that RPGD provides a practical and automatic solution for reliable extrinsic calibration of large-scale 3D HPE dataset collection.
\end{abstract}


\bstctlcite{IEEEexample:BSTcontrol}

\section{Introduction}
\label{sec:intro}

3D Human Pose Estimation (HPE) is a fundamental problem in computer vision spanned in fields including, but not limited to humanoid robotics, Human-Robot Interaction, Augmented and Virtual Reality, sports analytics, and digital humanities especially dance digitization. Recent advances in deep learning have significantly improved the pose estimation accuracy; however, these methods remain highly dependent on the availability of large-scale, accurately annotated 3D datasets. In contrast to 2D pose estimation, where annotations can be manually obtained from images, reliable 3D ground truth typically requires motion capture (MoCap) systems synchronized and calibrated with RGB cameras.

A critical challenge in collecting such datasets is extrinsic calibration—the accurate alignment between the coordinate system of the MoCap sensors and that of the RGB cameras. While intrinsic camera calibration and internal MoCap calibration are well-studied, estimating the rigid transformation between these heterogeneous sensing systems remains non-trivial. In practice, 3D HPE data collection is typically conducted in professional studios, where accurate extrinsic calibration is achievable. However, the high cost of such setups with fixed venue limit both scalability and accessibility of these datasets.

Existing calibration pipelines commonly rely on static calibration objects with rigid objects or patterns to establish accurate 3D–2D correspondences. Unfortunately, these assumptions break down in many realistic capture settings. When calibration is performed directly from human motion, both the 3D skeletal data---often affected by drift or sensor noise---and the 2D pose keypoints—frequently corrupted by occlusions, motion blur, or detector failures---introduce non-Gaussian noise and outliers. Classical geometric solvers such as the Perspective-n-Point (PnP) solver \cite{fischler1981random, 1217599} are highly sensitive to these imperfections, while purely iterative optimization methods like bundle adjustment are computationally expensive and sensitive to initialization, and become memory-intensive and slow for very large datasets.

These challenges motivate the need for a human-centric extrinsic calibration framework that can robustly operate under noisy and inconsistent conditions in real-world human pose data. In this work, we argue that human motion itself—when observed over long temporal sequences from monocular or multiple viewpoints—can serve as a powerful dynamic calibration target, provided that robustness and precision are jointly addressed. This paper presents \textbf{RPGD}, a robust and automatic extrinsic calibration framework aligning MoCap–based 3D human pose data with RGB videos. Compared to conventional PnP-based calibration pipelines, \textbf{RPGD} is explicitly designed to handle more noise, outliers, and temporal inconsistencies inherent in human pose annotations.

Our main contributions are threefold:
First, we formulate extrinsic calibration between MoCap systems and monocular or multi-view cameras as a human-pose-driven problem, enabling calibration directly from natural human motion without dedicated rigid calibration objects or likewise.
Second, we introduce a coarse-to-fine optimization strategy that tightly integrates RANSAC-P3P \cite{fischler1981random, 1217599} hypothesis generation with an analytic gradient-descent-based reprojection refinement, achieving both global robustness and sub-pixel accuracy on long motion sequences.
Third, we provide a comprehensive empirical study across three large-scale public 3D HPE datasets MPI-INF-3DHP\cite{mono-3dhp2017}, Human3.6M\cite{h36m_pami}, AIST++\cite{li2021learn}, and an in-the-wild capture setup, demonstrating that \textbf{RPGD} consistently recovers extrinsic parameters with accuracy comparable to the original ground truth.

\section{Related Work}

Extrinsic calibration between heterogeneous sensing systems has been a non-trivial issue in computer vision and robotics. However, calibrating RGB cameras directly from articulated human motion remains fundamentally different from classical calibration settings. Unlike rigid objects or calibration targets, human poses are high-dimensional, non-rigid, and subject to substantial measurement noise in both 2D and 3D. As a result, many established calibration and pose estimation techniques, which assume reliable correspondences and low-noise observations, do not directly transfer to human-centric capture scenarios.

Existing research relevant to this problem can be broadly grouped into two lines of work: First, 3D human pose estimation methods that depend on accurate extrinsic calibration; Second, extrinsic calibration approaches between MoCap systems and cameras, often relying on controlled setups or auxiliary objects, and geometric PnP-based solvers. While each of these areas has made significant progress independently, their intersection—robust extrinsic calibration from noisy human pose data—remains comparatively underexplored. Our work bridges this gap by explicitly designing a calibration framework that leverages the structure and temporal redundancy of human motion while remaining robust to annotation noise and outliers.

\subsection{3D Human Pose Estimation}
The field of 3D HPE has advanced rapidly, driven by the development of data collection sensors, large-scale annotated datasets, and excellent algorithms. More recently, Transformer-based architectures like ViTPose \cite{xu2022vitpose} and ViTPose++ \cite{10308645} have set new benchmarks by leveraging global attention mechanisms. However, the reliance on fully supervised data remains a bottleneck. Several works have explored unsupervised or semi-supervised paradigms. For instance, Pavllo et al.  \cite{pavllo20193d} proposed using temporal convolutions with semi-supervised training to lift 2D poses to 3D effectively. Early deep learning approaches relied on direct regression from images, while recent methods have exploited spatial or temporal information to resolve depth ambiguities. Techniques include manifold-constrained multi-hypothesis estimation \cite{rommel2024manipose}, non-rigid structure-from-motion modeling \cite{ji2024unsupervised}, and geometry-consistent loss functions \cite{matsune2024geometry} to enforce anatomical validity. Pseudo-labeling strategies have also been employed to fine-tune models on domain-specific data, such as sports motions \cite{suzuki2024pseudo}. Comprehensive surveys \cite{neupane2024survey} highlight that despite these algorithmic advances, the performance upper bound is still dictated by the limitation of the 3D annotations and diversity, which necessitates precise calibration between the visual and 3D metric tracking systems.

\subsection{Extrinsic Matrix Calibration }

Traditionally, sensor calibration relies on static features or calibration rigs. However, recent works have demonstrated that human subjects can serve as dynamic calibration cues. Lee et al. \cite{lee2022extrinsic} demonstrated that a moving person can serve as the calibration target, using keypoints detected from the human body to solve for extrinsics. Xu et al. \cite{xu2021wide} proposed utilizing person re-identification cues to solve wide-baseline calibration. Building on the utility of human subjects, Ma et al. \cite{ma2022virtual} introduced 'virtual correspondence', leveraging human pose to establish geometry in extreme views where standard feature matching fails. More recently, Müller et al. \cite{muller2025reconstructing} advanced this direction by jointly reconstructing people, static environments, and camera parameters, demonstrating that dynamic human motion provides sufficient signal for full scene and camera recovery.  Several studies also have focused on specific setups, such as calibrating wearable exoskeletons \cite{mileti2025extrinsic} or using prior knowledge of human motion to constrain the solution space \cite{moliner2021better}. 
Unlike the above works, which focus on self-calibration within a camera-based infrastructure (solving for relative camera poses) or motion tracking sensors, our approach explicitly registers RGB sensors to an external MoCap reference frame.

\section{Method}

Calibrating the extrinsic parameters between a MoCap system and RGB cameras is a classic yet persistent challenge. Unlike standard pure stereo camera calibration, aligning a MoCap frame to a camera frame requires finding correspondences between 3D features and relative 2D image keypoints. While classical PnP solvers like the minimal P3P solver \cite{1217599} provide efficient closed-form solutions, they are highly sensitive to noisy correspondences---a common issue when aligning MoCap data with visual pose detections. Our work builds upon these solvers but integrates them into a robust optimization framework to handle the noise inherent in human pose data.

\begin{figure}[t]
    \centering
    \includegraphics[
        width=0.95\columnwidth, 
        trim=0 120 0 40,      
        clip
    ]{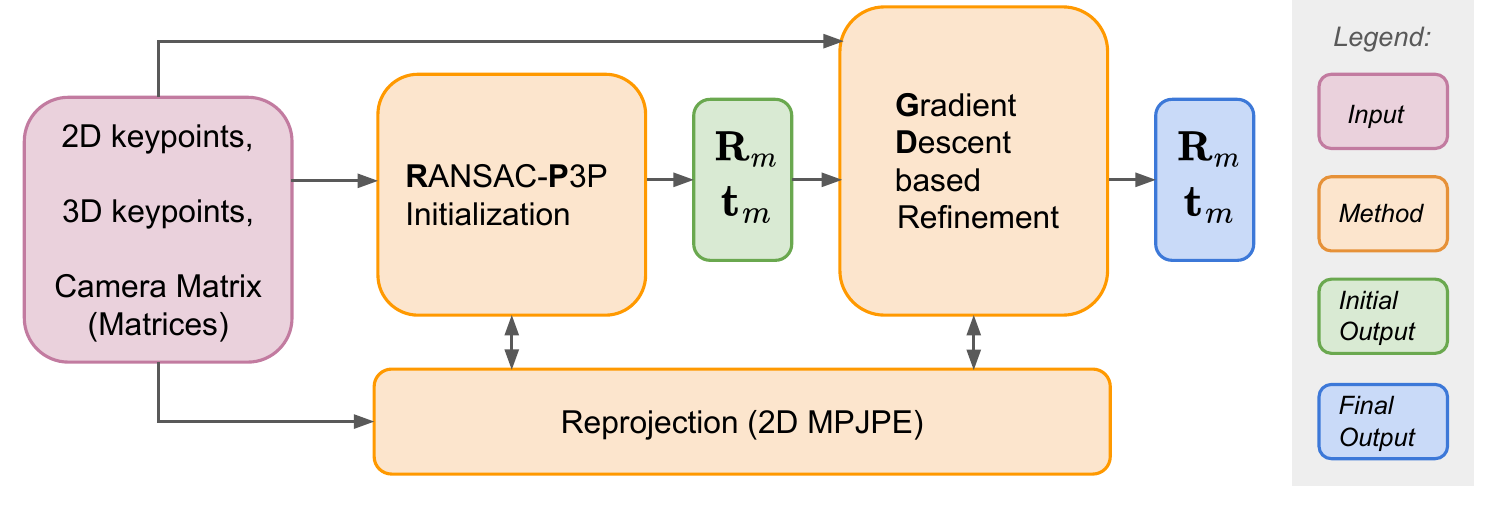}
    \caption{Overview of the \textbf{RPGD} extrinsic calibration framework. The pipeline consists of two stages: First, a RANSAC-P3P solver estimates the initial rotation $\mathbf{R}_m$ and translation $\mathbf{t}_m$ by maximizing the number of inliers. Second, the estimation from step one is fine-tuned based on gradient descent to minimize the reprojection error of 2D MPJPE.}
    \label{fig:pipeline}
\end{figure}

\subsection{Problem Formulation}

We consider the problem as estimating the rigid transformation that aligns 3D human joint positions obtained from a MoCap system with their corresponding 2D keypoints from one or more RGB cameras. We assume there is only one person in the scene, multiple person re-identification and matching across views is another topic beyond. Let $N$ be the number of RGB cameras and $J$ be the number of human body joints, and the time frames are indexed by $t \in \{1, \dots, T\}$. 
The multi-view system consists of $N$ synchronized cameras modeled by the standard pinhole model. For the $i$-th camera, let $\mathbf{P}_i \in \mathbb{R}^{3\times4}$ denote the camera matrix, defined as $\mathbf{P}_i = \mathbf{K}_i [\mathbf{R}_{c,i} \mid \mathbf{t}_{c,i}]$, where $\mathbf{K}_i\in \mathbb{R}^{3\times3}$ is the intrinsic matrix, and $\mathbf{R}_{c,i}, \mathbf{t}_{c,i}$ represent
the camera extrinsic matrix from the world coordinates to the camera coordinates. Lens distortion is considered when parameters are available.

Let $\mathbf{W}_{jt} \in \mathbb{R}^{3}$ denote the 3D position of joint $j$ at time $t$ in the MoCap coordinate system. Let $\mathbf{w}_{ijt} \in \mathbb{R}^{2}$ be the labeled 2D keypoints of this joint $j$ in camera $i$ at time $t$. Our goal is to estimate the rigid body transformation, represented by a rotation matrix $\mathbf{R}_m \in \mathbb{R}^{3\times3}$ and a translation vector $\mathbf{t}_m \in \mathbb{R}^{3}$, which transform points from the MoCap coordinate system to the world coordinate system.

The projection of a 3D joint $\mathbf{W}_{jt}$ onto the image plane of camera $i$ at time $t$ is given by:
\begin{equation}
\lambda \begin{bmatrix} \mathbf{u}_{ijt} \\ 1 \end{bmatrix} = \mathbf{P}_i 
\begin{bmatrix} \mathbf{R}_m & \mathbf{t}_m \\ \mathbf{0}^\top & 1 \end{bmatrix}
\begin{bmatrix} \mathbf{W}_{jt} \\ 1 \end{bmatrix},
\label{eq:projection}
\end{equation}
where $\mathbf{u}_{ijt} \in \mathbb{R}^2$ is the projected 2D keypoint and $\lambda$ is the depth scalar.

To estimate $(\mathbf{R}_m, \mathbf{t}_m)$, we adopt a coarse-to-fine optimization strategy. The proposed \textbf{RPGD} framework decomposes the estimation into two complementary stages: a robust geometric initialization that provides global consistency, followed by a refinement stage that exploits temporal redundancy for high-precision alignment. This design explicitly balances robustness and accuracy, two competing objectives that are difficult to achieve simultaneously with a single optimization procedure.

As illustrated in Fig.~\ref{fig:pipeline}, the \textbf{RPGD} pipeline takes synchronized 2D and 3D keypoints $\mathbf{w}_{ijt}$ and $\mathbf{W}_{jt}$  as well as camera matrix $\mathbf{P}_i$ as the input. The process consists of two main stages:
\begin{enumerate}
    \item RANSAC-P3P Initialization: By randomly sampling minimal subsets of 3D-2D keypoint correspondences, we generate candidate poses and select the one that maximizes the number of inliers.
    \item Gradient-Descent-based Refinement: The best hypothesis from the initialization stage serves as the starting point. Here, we minimize the aggregate reprojection error over all valid keypoints based on gradient descent algorithm. 
\end{enumerate}
The feedback loop shown in Fig.~\ref{fig:pipeline} highlights that the reprojection error drives both the selection of the best geometric hypothesis and the final parameter refinement.

\subsection{RANSAC-P3P Initialization}
In the first stage, we obtain a robust initial estimate of the extrinsic parameters by casting the calibration task as a PnP problem where n is 3, and solving it using a RANSAC strategy \cite{fischler1981random} coupled with a P3P solver \cite{1217599}. Although human joints are non-rigid, each joint correspondence at a given time instant provides a valid 3D–2D constraint. By sampling minimal subsets of such correspondences across time and views, we generate candidate hypotheses while explicitly tolerating outliers caused by occlusions, noise, and so on.

The whole procedure of RANSAC-P3P is detailed in Algorithm~\ref{alg:ransac_p3p}. In each RANSAC iteration, we randomly sample 3 correspondences (triplets of $(i, j, t)$) with same $i$ and $t$ from the entire sequence. We calculate candidate poses $(\mathbf{R}_m, \mathbf{t}_m)$ and verify them against all available pairs of 2D and 3D keypoints $\mathbf{w}_{ijt}$ and $\mathbf{W}_{jt}$.  We obtain the projected 2D keypoints $\mathbf{u}_{ijt}$ by \eqref{eq:projection}. Let $\mathbf{r}_{ijt} \in \mathbb{R}^{2}$ be the residual vector in pixel: 
\begin{equation}
    \mathbf{r}_{ijt} = \mathbf{u}_{ijt} - \mathbf{w}_{ijt}.
    \label{eq:r}
\end{equation}

A projection is considered as an inlier if the $L2$ norm of  $\mathbf{r}_{ijt}$ falls below a threshold $\tau$:
\begin{equation}
    \| \mathbf{r}_{ijt}  \|_2 < \tau.
\end{equation}
The hypothesis with the maximum inlier count is selected as the initial estimation of $(\mathbf{R}_m, \mathbf{t}_m)$. Since 2D and 3D keypoint annotations always contain noise or outliers, direct optimization is prone to local minima, especially for a very long sequence. Downsampling is applied to reduce the risk of resulting in local minima.

\begin{algorithm}[t]
\caption{RANSAC-P3P Initialization}
\label{alg:ransac_p3p}
\begin{algorithmic}[1]

\REQUIRE Camera  matrices $\{\mathbf{P}_i\}_{i=1}^{N}$,
3D keypoints $\{\mathbf{W}_{jt}\}$,
2D keypoints $\{\mathbf{w}_{ijt}\}$,
inlier threshold $\tau$,
number of RANSAC iterations $K$

\ENSURE Initial extrinsic parameters $(\mathbf{R}_m^{}, \mathbf{t}_m^{})$

\STATE Initialize best inlier set $\mathcal{I}^{*} \leftarrow \emptyset$

\STATE Construct correspondence set
\[
\mathcal{C} = \{(i,j,t)\;|\; \mathbf{w}_{ijt} \text{ is valid}\}
\]

\FOR{$k = 1$ to $K$}
    \STATE Randomly sample three distinct correspondences from the same camera view and time
    \[
    \{(i,j_1,t),(i,j_2,t),(i,j_3,t)\} \subset \mathcal{C}
    \]
    \STATE Compute candidate pose solutions
    \[
    \{(\mathbf{R}_m^{k,l}, \mathbf{t}_m^{k,l})\}_{l=1}^{L}
    \]
    using a P3P solver

    \FOR{each candidate $(\mathbf{R}_m^{k,l}, \mathbf{t}_m^{k,l})$}
        \STATE Initialize inlier set $\mathcal{I}_{k,l} \leftarrow \emptyset$

        \FOR{each correspondence $(i,j,t) \in \mathcal{C}$}
            \STATE Project $\mathbf{W}_{jt}$ into camera $i$ by \eqref{eq:projection}
            \IF{depth $\lambda \leq 0$}
                \STATE \textbf{continue}
            \ENDIF
            \STATE Compute reprojection residual $\mathbf{r}_{ijt}$ by \eqref{eq:r}

            \IF{$\|\mathbf{r}_{ijt}\|_2 < \tau$}
                \STATE $\mathcal{I}_{k,l} \leftarrow \mathcal{I}_{k,l} \cup \{(i,j,t)\}$
            \ENDIF
        \ENDFOR

        \IF{$|\mathcal{I}_{k,l}| > |\mathcal{I}^{*}|$}
            \STATE $\mathcal{I}^{*} \leftarrow \mathcal{I}_{k,l}$
            \STATE $(\mathbf{R}_m^{}, \mathbf{t}_m^{}) \leftarrow (\mathbf{R}_m^{k,l}, \mathbf{t}_m^{k,l})$
        \ENDIF
    \ENDFOR
\ENDFOR

\RETURN $(\mathbf{R}_m^{}, \mathbf{t}_m^{})$

\end{algorithmic}
\end{algorithm}

\subsection{Gradient-Descent-based Refinement}
While RANSAC-P3P provides robustness to gross outliers, its discrete hypothesis selection limits the achievable accuracy, particularly in the presence of moderate but systematic annotation noise. We therefore refine the initial estimate by minimizing the reprojection error over all valid correspondences using gradient-descent-based optimization. This second stage exploits a large number of observations available across time frames and cameras, enabling precise alignment through continuous optimization. 

We parameterize the rigid transformation using a 3D translation vector $\mathbf{t}_m$ as before and Z$(\gamma)$Y$(\beta)$X$(\alpha)$ Euler angles for rotation matrix $\mathbf{R}_m$. Although alternative representations such as quaternion or Lie algebra parameterizations are possible, Euler angles provide a compact and computationally efficient formulation in our setting, where the rotation range is limited by the RANSAC initialization.

\subsubsection{Objective Function}
We minimize the loss $\mathcal{L}$ defined below:
\begin{equation}
    \mathcal{L}(\alpha, \beta, \gamma, \mathbf{t}_m) = \frac{1}{2|\mathcal{S}|} \sum_{(i,j,t) \in \mathcal{S}} \| \mathbf{r}_{ijt} \|_2^2,
    \label{eq:loss}
\end{equation}
where $\mathcal{S}$ is the set of all correspondences of one sequence with positive depth, i.e., $\lambda > 0$.

\subsubsection{Gradient Computation}
We utilize the chain rule to propagate gradients. The gradient with respect to the translation $\mathbf{t}_m$ is:
\begin{equation}
    \frac{\partial \mathcal{L}}{\partial \mathbf{t}_m} = \frac{1}{|\mathcal{S}|} \sum_{(i,j,t)\in \mathcal{S}} \mathbf{J}_{i}^\top \mathbf{r}_{ijt},
\end{equation}
where $\mathbf{J}_{i}\in \mathbb{R}^{2\times3}$ is the Jacobian matrix of the projection function \eqref{eq:projection} from  $\mathbf{W}_{jt}$ to $\mathbf{u}_{ijt}$ in the coordinates of camera $i$ . Similarly, the gradients for the Euler angles are derived via the chain rule on the rotation matrix $\mathbf{R}_m(\alpha, \beta, \gamma)$. For an angle $\theta \in \{\alpha, \beta, \gamma\}$:
\begin{equation}
    \frac{\partial \mathcal{L}}{\partial \theta} = \frac{1}{|\mathcal{S}|} \sum_{(i,j,t)\in \mathcal{S}} \mathbf{r}_{ijt}^\top  \mathbf{J}_{i} \mathbf{R}_{c,i} \frac{\partial \mathbf{R}_m}{\partial \theta} \mathbf{W}_{jt}.
\end{equation}

\subsubsection{Optimization Details}

We optimize the loss $\mathcal{L}$ using the Adam optimizer \cite{kingma2014adam} with separate learning rates for rotation and translation parameters to account for their different numerical scales. A cosine annealing schedule is employed to stabilize convergence and reduce oscillations. To further improve robustness, optimization is performed on a temporally downsampled subset of frames, which reduces overfitting to noisy measurements while preserving global geometric consistency.

\section{Experiments}

We evaluate the proposed \textbf{RPGD} method on three different public datasets: MPI-INF-3DHP \cite{mono-3dhp2017}, Human3.6M \cite{h36m_pami}, and AIST++ \cite{li2021learn}, covering a wide range of motion capture types, sequence length, keypoints numbers, and average Ground Truth (GT) of 2D Mean Per Joint Position Error (MPJPE) accuracy defined in \eqref{eq:mpjpe}. Additionally, we conduct a qualitative evaluation on a self-collected dataset to demonstrate generalization to in-the-wild scenarios.

\subsection{Datasets}
Table~\ref{tab:dataset_comparison} summarizes the key characteristics of the three public datasets we evaluated on, in terms of the number of frames, number of subjects, number of cameras, number of frame rate per second (FPS), number of keypoints, ground-truth 2D MPJPE at the pixel level, camera matrix availability, distortion parameters availability, whether physical 3D MoCap system is involved, and whether 2D annotations are projected by 3D keypoints or not.
\begin{itemize}
    \item MPI-INF-3DHP \cite{mono-3dhp2017}: A dataset captured in a controlled studio environment with 14 cameras. It features 8 subjects performing 8 activity sets. Note that for this dataset, 3D annotations are provided in the camera coordinate system. 2D keypoints annotation is by the projection of 3D annotation, resulting the average GT 2D MPJPE of $10^{-1}$ pixel level.
    \item Human3.6M \cite{h36m_pami}: A large-scale indoor dataset including 3.6 million frames. We utilize the training set of 7 subjects (S1, S5, S6, S7, S8, S9, S11) across all 30 action sequences. 3D keypoints annotation is captured by 10 motion cameras, the average GT 2D MPJPE is of $10^0$ pixel level. 
    \item AIST++ \cite{li2021learn}: A dance motion dataset derived from the AIST Dance Video Database \cite{tsuchida2019aist}, containing 10.1 million frames of 30 subjects performing 10 different genres of dances. The 3D keypoint annotations of this dataset is not from a physical MoCap system, having an average GT 2D MPJPE of $10^1$ pixel level. 
\end{itemize}

\begin{table}[htbp]
\centering
\caption{Features of Three Public 3D HPE Datasets.}
\label{tab:dataset_comparison}
\small
\renewcommand{\arraystretch}{1.0}
\begin{tabularx}
{\columnwidth}{Xccc} 
\toprule
Feature & MPI-INF-3DHP & Human3.6M & AIST++ \\ \midrule
\# Frames & 1.3M & 3.6M & 10.1M \\
\# Subjects & 8 & 11 & 30 \\
\# RGB Cameras & 14 & 4 & 9 \\
\# FPS & 25/50 & 50 & 60 \\
\# Keypoints & 28 & 32 & 17 \\
2D MPJPE (GT) & $10^{-1}$ & $10^{0}$ & $10^{1}$ \\
Camera matrix & \cmark & \cmark & \cmark \\
Distortion & \xmark & \cmark & \cmark \\
Physical MoCap & \cmark & \cmark & \xmark \\
Non-projected 2D & \xmark & \cmark & \cmark \\
\bottomrule
\end{tabularx}
\end{table}

\subsection{Implementation Details}
For the RANSAC-P3P, we implement the P3P solver using the OpenCV Library \cite{bradski2000opencv}, adapted to handle keypoints in different coordinate systems. To ensure efficiency and robustness, we employ a temporal downsampling strategy. Specifically, we sample frames at a lower frequency for the RANSAC-P3P to improve the efficiency.

For the gradient-descent-based refinement, we use a denser downsampling rate to reduce the risk of  resulting in local minima, especially for long sequences. On the MPI-INF-3DHP dataset, 2 sequences with abnormal 2D MPJPE outliers are excluded, and 14 sequences are evaluated. On the Human3.6M dataset, we evaluate through the whole dataset. On the AIST++ dataset, we follow the official protocol, excluding 45 sequences labeled as unreliable, and finally evaluate on 826 sequences filtered by a maximum 2D MPJPE of 300 pixels.

\subsection{Evaluation Protocol}
The primary metric for evaluation is the 2D MPJPE in pixel, defined as the mean Euclidean distance of pixel between the 2D points projected from 3D keypoints and the ground truth 2D annotations: 
\begin{equation}
    \text{MPJPE} = \frac{1}{|\mathcal{S}|} \sum_{(i,j,t) \in \mathcal{S}} \| \mathbf{r}_{ijt}  \|_2.
    \label{eq:mpjpe}
\end{equation}
We compare the 2D MPJPE of three setups:
\begin{enumerate}
    \item Ground Truth (GT): The original error using the provided ground truth extrinsic parameters from the dataset itself. This serves as the theoretical lower bound.
    \item RANSAC-P3P: The initial output after the premier step by RANSAC-P3P.
    \item \textbf{RPGD} (Ours): The final result combining RANSAC-P3P and gradient-descent-based refinement.
\end{enumerate}

\section{Results}

\subsection{Quantitative Results}

\begin{figure*}[t]
    \centering
    
    % --- ROW 1: Two plots side-by-side ---
    \begin{minipage}[b]{0.2\textwidth}
        \centering
        % Adjust 'height' to control aspect ratio (e.g., 5cm)
        \includegraphics[width=\linewidth, height=5cm, keepaspectratio]{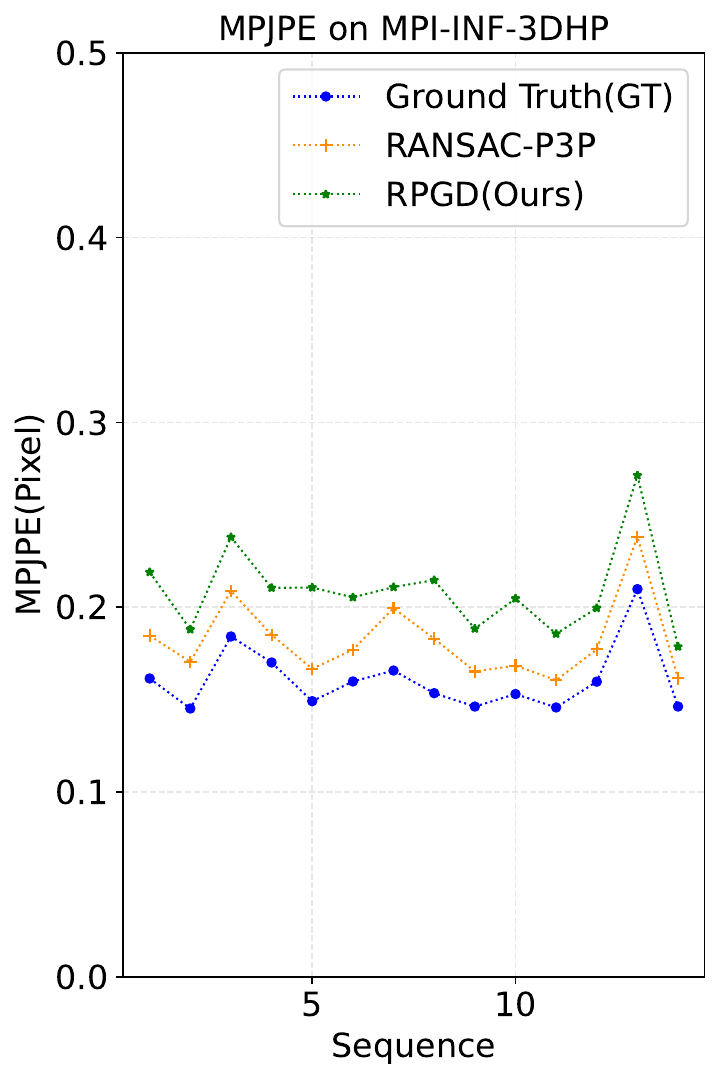}
        \captionof*{figure}{\makebox[\textwidth][l]{        (a) MPJPE on MPI-INF-3DHP}}
      
    \end{minipage}
    \hfill % Pushes the two plots to the edges
    \begin{minipage}[b]{0.78\textwidth}
        \centering
        \includegraphics[width=\linewidth, height=5cm, keepaspectratio]{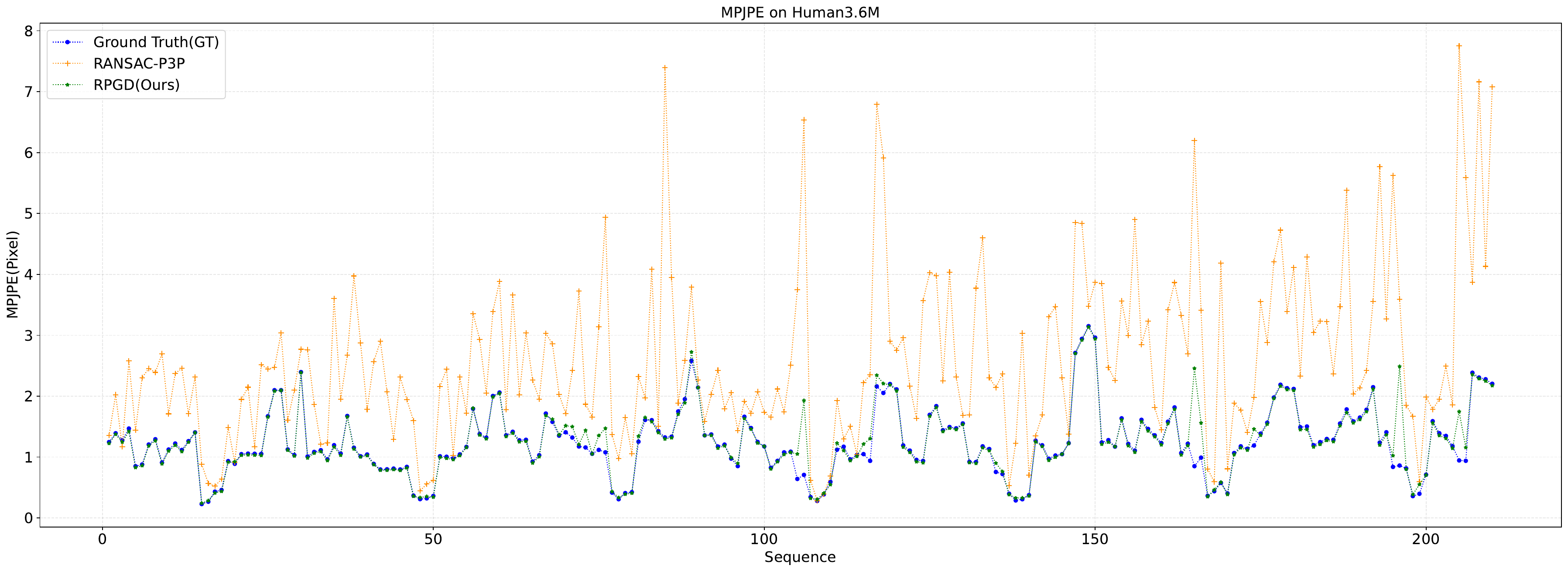}
        \captionof*{figure}{(b) MPJPE on Human3.6M }
    \end{minipage}
    
    %\vspace{0.1em} % Vertical space between rows
    
    % --- ROW 2: One plot centered ---
    \begin{minipage}[b]{1\textwidth}
        \centering
        \includegraphics[width=\textwidth, height=5cm, keepaspectratio]{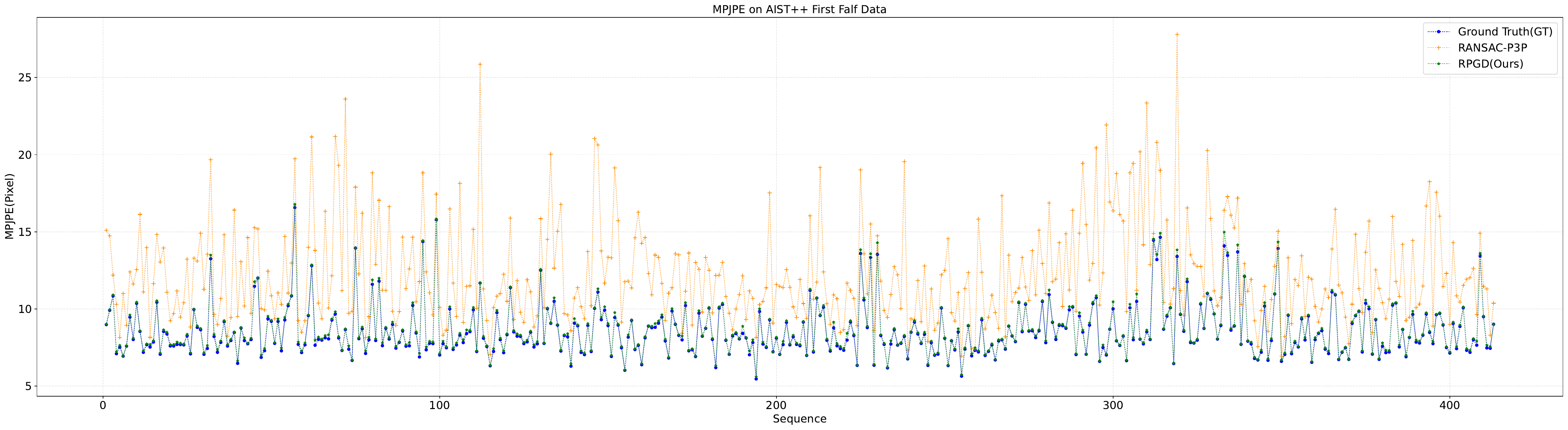}
        \captionof*{figure}{(c-1) MPJPE on First Half Sequences of AIST++}
    \end{minipage}

        % --- ROW 3: One plot centered ---
    \begin{minipage}[b]{1\textwidth}
        \centering
        \includegraphics[width=\textwidth, height=8cm, keepaspectratio]{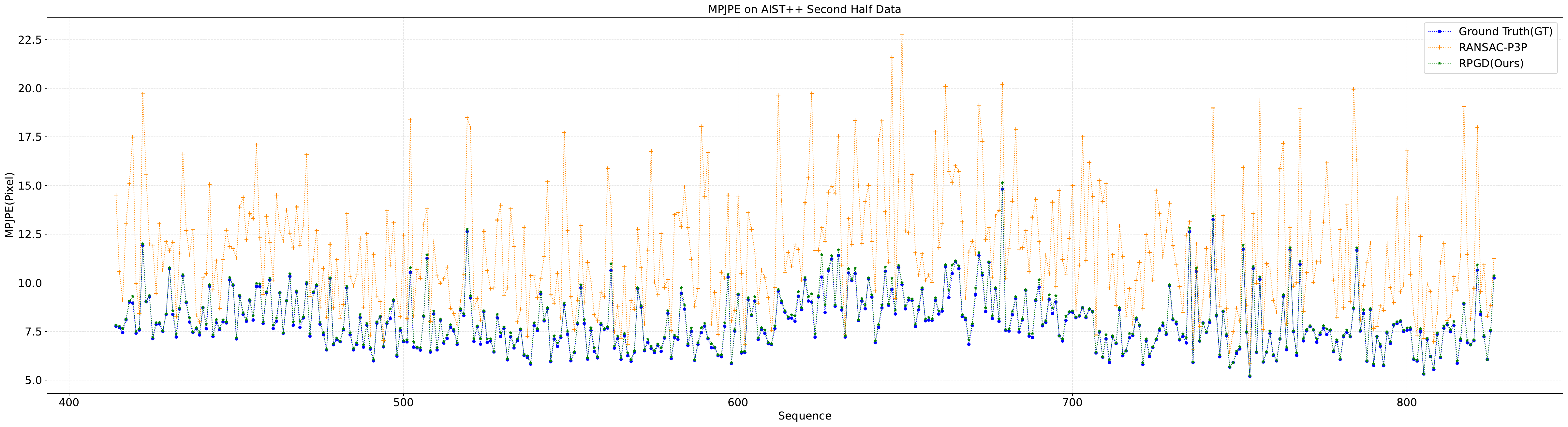}
        \captionof*{figure}{(c-2) MPJPE on Second Half Sequences of AIST++}
    \end{minipage}

    % --- Main Caption ---
    \caption{
    Average 2D MPJPE (in Pixel) of each sequence across three public 3D HPE datasets. The rounds in blue indicate the average GT MPJPE , the plus in orange indicate the average MPJPE after RANSAC-P3P, the stars in green indicate the average MPJPE by \textbf{RPGD}. The X-axis represents the Sequence Index, the Y-axis is the MPJPE value in pixel. 
\textbf{(a)} MPJPE on each MPI-INF-3DHP Sequence. 
\textbf{(b)} MPJPE on each Human3.6M Sequence. \textbf{(c-1, c-2)} MPJPE on each AIST++ Sequence.}
    \label{fig:quantitative_plots}
\end{figure*}

Table~\ref{tab:dataset_MPJPE} presents the quantitative comparison of the average 2D MPJPE (pixels) of all the sequences in three datasets.

\subsubsection{MPI-INF-3DHP} On this dataset, the RANSAC-P3P achieves near-perfect average accuracy (0.18 pixels), marginally outperforming the RPGD result (0.21 pixels). This result is expected given the synthetic nature of the 2D annotations of MPI-INF-3DHP, which are generated via direct projection and thus lack the measurement noise typical of real-world detectors or human annotation. Crucially, however, the error for both stages remains well below 0.3 pixel on average, confirming that \textbf{RPGD} maintains high-fidelity calibration.

\subsubsection{Human3.6M} Our method achieves significant improvement on Human3.6M dataset. On average, the RANSAC-P3P initialization yields an error of 2.59 pixels, which is refined by \textbf{RPGD} to 1.26 pixels—nearly matching the ground truth error of 1.23 pixels. This result demonstrates the efficacy of gradient-descent-based refinement in fine-tuning the result.

\subsubsection{AIST++} The improvement is more pronounced on AIST++ dataset. RANSAC-P3P results in a high error of $\sim$12 pixels. \textbf{RPGD} successfully reduces this to 8.39 pixels on average, recovering the extrinsic matrix with high accuracy comparable to the ground truth (8.28 pixels).

\begin{table}[htbp]
\centering
\caption{Average 2D MPJPE (pixel) of GT, RANSAC-P3P and \textbf{RPGD} over three public 3D HPE Datasets. GT represents the ground truth MPJPE of the original dataset.}
\label{tab:dataset_MPJPE}
\small % IEEE standard table font size
\renewcommand{\arraystretch}{1.0} % standard row spacing
\begin{tabular}{lccc}
\toprule
MPJPE (pixel) $\downarrow$  & MPI-INF-3DHP & Human3.6M & AIST++ \\ \midrule
GT & 0.1607 & 1.2299 & 8.2837 \\ \midrule
RANSAC-P3P  & \textbf{0.1819} & 2.5906 & 11.9826 \\
\textbf{RPGD} (Ours) & 0.2089 & \textbf{1.2605} & \textbf{8.3916} \\
\bottomrule
\end{tabular}
\end{table}

We observe a distinct contrast between the results on MPI-INF-3DHP and the other datasets. On MPI-INF-3DHP dataset, the 2D annotations are the projections of the 3D keypoints, resulting in near-zero noise. The RANSAC-P3P initialization alone achieves sub-pixel accuracy (0.18 pixel), and the gradient-descent-based refinement does not optimize the result. However, on Human3.6M and AIST++ datasets, where 2D keypoints are generated by extertial detectors or manual annotation containing real-world noise or outliers, RANSAC-P3P leaves a significant residual error (2.59 pixels and 11.98 pixels). Here, the proposed \textbf{RPGD} refinement is crucial, reducing error by roughly 50\% and 30\% respectively. This result confirms that while geometric solvers suffice for synthetic data, the proposed gradient-descent-based refinement is essential for accurate calibration in practical, noisy environments.

As visualized in Fig.~\ref{fig:quantitative_plots}, we report the 2D MPJPE for each sequence in each dataset. On the datasets utilizing non-projected 2D keypoints (Human3.6M and AIST++ in Fig.~\ref{fig:quantitative_plots} (b), (c-1), (c-2)), the RANSAC-P3P initialization exhibits high variance across different sequences. This fluctuation is expected, as non-projected annotations inherently contain varying levels of inconsistency like noise or outliers. In these scenarios, the gradient-descent-based refinement stage consistently reduces the MPJPE, pulling the result closer to the ground truth baseline for nearly every sequence. Conversely, on the projected 2D annotation within MPI-INF-3DHP (Fig.~\ref{fig:quantitative_plots} (a)), where the geometric relationship is ideal, both RANSAC-P3P and \textbf{RPGD} achieve consistently low error across all sequences.

\subsection{Qualitative Evaluation}
To validate the generalization capability of \textbf{RPGD} beyond standard benchmarks, we conducted an experiment in an uncontrolled 'in-the-wild' environment. We utilized a Microsoft Kinect Xbox 360 sensor \cite{6190806, kinect-openni-bvh-saver, meshonline_mocap} to capture 3D skeletal keypoints and a separate RGB camera to record videos. The RGB camera's intrinsic matrix is calibrated with a chessboard by OpenCV Library\cite{bradski2000opencv}.

\begin{figure}[ht]
    \centering
    
    \includegraphics[width=0.95\columnwidth, trim=0 100 0 5, clip]{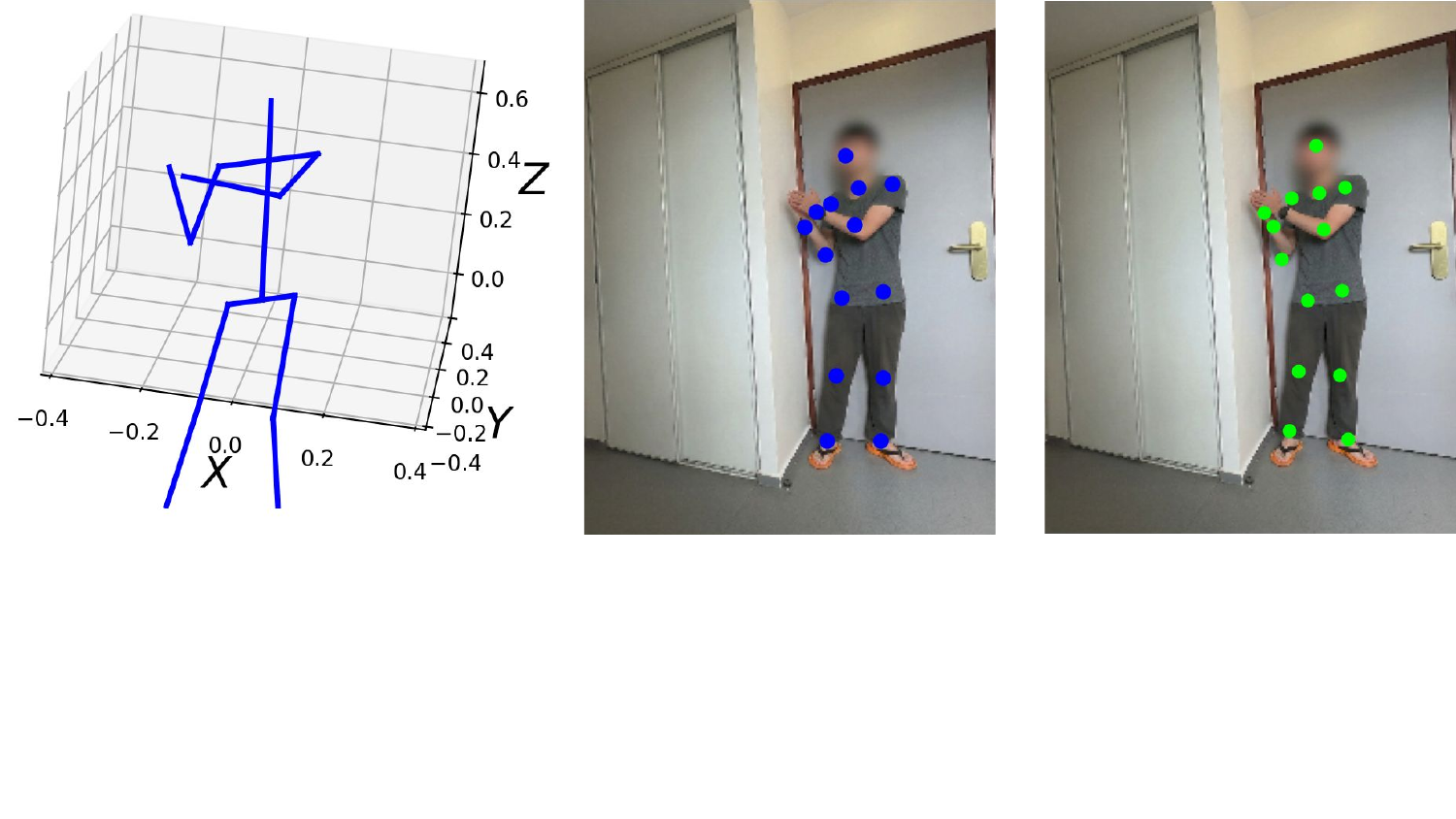}
    \caption{Qualitative evaluation on in-the-wild scene. \textbf{Left:} Raw 3D skeleton keypoints captured by Kinect sensor. \textbf{Middle:} Ground truth 2D keypoints annotation. \textbf{Right:} 2D keypoints projected from raw 3D skeleton keypoints by extrinsic matrix from \textbf{RPGD}.}
    \label{fig:inthewild}
\end{figure}

As shown in Fig.~\ref{fig:inthewild}, the raw 3D data from Kinect (Left) and the image from the RGB camera (Middle) are initially in different coordinate systems. By applying \textbf{RPGD}, we recover the extrinsic matrix to align the Kinect's metric frame with the camera's visual frame. The result is visualized in the right panel of Fig.~\ref{fig:inthewild}, where the reprojected 3D skeleton keypoints overlay the subject. Despite the lack of calibration in advance between the MoCap system and the RGB camera, our method successfully recovers the spatial relationship, confirming its potential application on extrinsic calibration for real-world 3D HPE dataset collection setups.

\section{Discussion}

Despite its robustness and practical effectiveness, the \textbf{RPGD} framework still has space for further improvement.

\subsubsection{Sensitivity to Intrinsics} While the RANSAC-P3P initialization provides robustness against minor calibration deviations by treating geometric inconsistencies as outliers, significant intrinsic errors  may inevitably degrade the convergence and reduce the final accuracy. 

\subsubsection{Multi-Person Extension}
For multi-person scenarios, identity association across time and views becomes the challenge. \textbf{RPGD} can be extended to these complex scenes by incorporating human tracking or re-identification modules. 

\subsubsection{Runtime Efficiency} Table~\ref{tab:dataset_runtime} shows that the average processing time per frame is on the millisecond level, making our method suitable for large-scale dataset batch processing. We believe it is still possible to reduce it by optimizing the low-level implementation.
\begin{table}[htbp]
\centering
\caption{Average precessing time per frame (in millisecond) over three public 3D HPE datasets by \textbf{RPGD}.}
\label{tab:dataset_runtime}
{\small % IEEE standard table font size
{\begin{tabular}{lccc}
\toprule
Dataset & MPI-INF-3DHP & Human3.6M & AIST++ \\ \midrule
Time per frame & 5.2 ms & 1.2 ms & 2.3 ms \\ 
\bottomrule
\end{tabular}
}}
\end{table}

\section{Conclusion}

We presented \textbf{RPGD}, a robust and automatic extrinsic calibration framework designed to align MoCap-based 3D human pose data with RGB camera videos. By explicitly addressing the noisy, outlier-prone 2D and 3D human pose keypoints, \textbf{RPGD} integrates RANSAC-P3P initialization with analytic gradient-descent-based refinement in a principled coarse-to-fine optimization scheme. Extensive experiments on three different public 3D HPE datasets and an in-the-wild setup demonstrate that \textbf{RPGD} consistently recovers extrinsic parameters with accuracy comparable to the original ground truth within sub-pixel 2D MPJPE level.

Beyond its immediate application to 3D HPE dataset construction, \textbf{RPGD} highlights the potential of human motion itself as a dynamic calibration target, without dedicated calibration objects or manual intervention. Our \textbf{RPGD} paradigm is particularly appealing for the extrinsic calibration of 3D HPE in unconstrained or less constrained environments.

\section*{Acknowledgment} 

This work is co-funded by the European Union. Views and opinions expressed are however those of the author(s) only and do not necessarily reflect those of the European Union or the granting authority. Neither the European Union nor the granting authority can be held responsible for them.

We thank the anonymous reviewers for their constructive comments and helpful suggestions.

\bibliographystyle{IEEEtran}
\bibliography{references} %\nocite{*}
\end{document}